\documentclass[runningheads]{llncs}

\usepackage{eccv}

\usepackage{eccvabbrv}

\usepackage{graphicx}
\usepackage{booktabs}

\usepackage[accsupp]{axessibility}  

\usepackage{xr-hyper}

\usepackage{hyperref}

\usepackage{orcidlink}

\usepackage{tabularx}
\usepackage{adjustbox}
\usepackage{float}
\usepackage{bm}
\usepackage{ifthen}
\usepackage{mathtools} 
\usepackage[capitalize]{cleveref} 

\definecolor{t1color}{HTML}{82368c}
\definecolor{t2color}{HTML}{e30613}
\definecolor{t3color}{HTML}{009640}

\newcommand{\tali}[2][]{%
    \ifthenelse{\equal{#1}{} }
        {\textcolor{magenta}{(TD) #2}}
        {\textcolor{magenta}{(TD) \sout{#1}\xspace{}#2}}
}

\newcommand{\narek}[2][]{%
    \ifthenelse{ \equal{#1}{} }
        {\textcolor{blue}{(NT) #2}}
        {\textcolor{blue}{(NT) \sout{#1}\xspace{}#2}}
}
\newcommand{\jonathon}[2][]{%
    \ifthenelse{ \equal{#1}{} }
        {\textcolor{cyan}{(JL) #2}}
        {\textcolor{cyan}{(JL) \sout{#1}\xspace{}#2}}
}
\newcommand{\samuel}[2][]{%
    \ifthenelse{ \equal{#1}{} }
        {\textcolor{orange}{(SRB) #2}}
        {\textcolor{orange}{(SRB) \sout{#1} #2\xspace{}}}
}
\newcommand{\denis}[2][]{%
    \ifthenelse{ \equal{#1}{} }
        {\textcolor{orange}{(DR) #2}}
        {\textcolor{orange}{(DR) \sout{#1} #2\xspace{}}}
}
\newcommand{\adam}[2][]{%
    \ifthenelse{ \equal{#1}{} }
        {\textcolor{purple}{(AH) #2}}
        {\textcolor{purple}{(AH) \sout{#1} #2\xspace{}}}
}
\newcommand{\peter}[2][]{%
    \ifthenelse{ \equal{#1}{} }
        {\textcolor{cyan}{(PK) #2}}
        {\textcolor{cyan}{(PK) \sout{#1} #2\xspace{}}}
}
\newcommand{\lorenzo}[2][]{%
    \ifthenelse{ \equal{#1}{} }
        {\textcolor{orange}{(LP) #2}}
        {\textcolor{orange}{(LP) \sout{#1} #2\xspace{}}}
}

\newcommand{\PAR}[1]{\noindent {\bf #1~}}

\newcommand{\boldparagraph}[1]{\vspace{4pt}\noindent{\bf #1} }    

\newcommand{\mathframe}{\mathbf{I}}
\newcommand{\mathmask}{\mathbf{M}}
\newcommand{\prescanframes}{\mathbf{I}_{\mathtt{S}}^r}
\newcommand{\prescanmasks}{\mathmask_{\mathtt{S}}^r}

\newcommand{\video}{\mathcal{I}_\text{video}}
\newcommand{\prescan}{\mathcal{I}_\text{scan}}
\newcommand{\masks}{\mathcal{M}}

\newcommand{\depth}{\mathbf{D}}

\newcommand{\bbox}{\mathcal{B}}

\newcommand{\gaussinit}{\mathcal{G}_\text{init}}
\newcommand{\gaussfg}{\mathcal{G}_\text{fg}}
\newcommand{\gausscanon}{\mathcal{G}_\text{can}}
\newcommand{\mugrid}{\bm{\mu}}

\newcommand{\cam}{\mathbf{T}}

\newcommand{\timestep}{t}
\newcommand{\rotquat}{\mathbf{Q}}

\newcommand{\translation}{\bm{\Delta}}

\newcommand{\posenc}{\gamma}
\newcommand{\focal}{f_\ell}

\newcommand{\point}{p}

\newcommand{\pcoarse}{\mathcal{G}_{\text{crs}}}
\newcommand{\nncoarse}{\mathcal{N}_{\text{crs}}}
\newcommand{\nn}{\mathcal{N}}

\newcommand{\mathcossim}{\text{cos-sim}}

\newcommand{\lossphoto}{\mathcal{L}_{\text{photo}}}
\newcommand{\losscanonref}{\mathcal{L}_{\text{can}}}
\newcommand{\losslone}{\mathcal{L}_{\text{L1}}}
\newcommand{\lossssim}{\mathcal{L}_{\text{SSIM}}}
\newcommand{\lambdassim}{\lambda_{\text{SSIM}}}
\newcommand{\losstv}{\mathcal{L}_{\text{TV}}}
\newcommand{\lambdatv}{\lambda_{\text{TV}}}

\newcommand{\lossdepth}{\mathcal{L}_{\text{depth}}}
\newcommand{\lambdadepth}{\lambda_{\text{depth}}}

\newcommand{\losstrack}{\mathcal{L}_{\text{track}}}
\newcommand{\lambdatrack}{\lambda_{\text{track}}}

\newcommand{\lossreproj}{\mathcal{L}_{\text{reproj}}}
\newcommand{\lambdareproj}{\lambda_{\text{reproj}}}

\newcommand{\losscoarseiso}{\mathcal{L}_{\text{crs\_iso}}}
\newcommand{\lambdacoarseiso}{\lambda_{\text{crs\_iso}}}
\newcommand{\lossdenseiso}{\mathcal{L}_{\text{dense\_iso}}}
\newcommand{\lambdadenseiso}{\lambda_{\text{dense\_iso}}}
\newcommand{\lossrigid}{\mathcal{L}_{\text{rigid}}}
\newcommand{\lambdarigid}{\lambda_{\text{rigid}}}

\newcommand{\motiondipmath}{\Phi_{\theta}}

\newcommand{\ourmethod}{DRoPS}
\newcommand{\motiondip}{Deep Motion Prior}
\newcommand{\motiondipshort}{DMP}

\makeatletter
\renewcommand\section{\@startsection{section}{1}{\z@}%
                       {-12\p@ \@plus -2\p@ \@minus -2\p@}%
                       {6\p@ \@plus 2\p@ \@minus 2\p@}%
                       {\normalfont\large\bfseries\boldmath
                        \rightskip=\z@ \@plus 8em\pretolerance=10000 }}
\renewcommand\subsection{\@startsection{subsection}{2}{\z@}%
                       {-10\p@ \@plus -2\p@ \@minus -2\p@}%
                       {4\p@ \@plus 2\p@ \@minus 2\p@}%
                       {\normalfont\normalsize\bfseries\boldmath
                        \rightskip=\z@ \@plus 8em\pretolerance=10000 }}
\renewcommand\subsubsection{\@startsection{subsubsection}{3}{\z@}%
                       {-8\p@ \@plus -2\p@ \@minus -2\p@}%
                       {-0.5em \@plus -0.22em \@minus -0.1em}%
                       {\normalfont\normalsize\bfseries\boldmath}}
\makeatother

\begin{document}

\title{DRoPS: \textbf{D}ynamic 3D \textbf{R}econstruction \textbf{o}f \textbf{P}re-\textbf{S}canned Objects}

\titlerunning{DRoPS: Dynamic 3D Reconstruction of Pre-Scanned Objects}

\author{Narek Tumanyan\inst{1,2} \and
Samuel Rota Bulò\inst{2} \and
Denis Rozumny\inst{2} \and
Lorenzo Porzi\inst{2} \and
Adam Harley\inst{2} \and
Tali Dekel\inst{1} \and
Peter Kontschieder\inst{2} \and
Jonathon Luiten\inst{2}}

\authorrunning{N.~Tumanyan et al.}

\institute{Weizmann Institute of Science \and
Meta Reality Labs \\
{\small Project webpage: \href{https://drops-dynamics.github.io}{drops-dynamics.github.io}}}

\maketitle

\begin{centering}
\captionsetup{hypcap=false}
\label{fig:teaser}
\includegraphics[width=\textwidth]{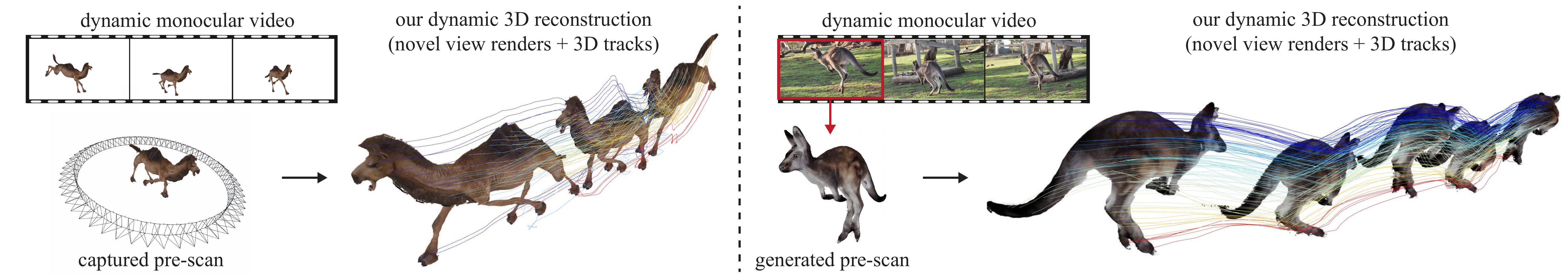}
\captionof{figure}{
  Given a static pre-scan and a monocular video of a dynamic object, \ourmethod{} reconstructs a complete dynamic 3D representation, enabling high-quality novel view synthesis. 
  The pre-scan is either captured (left) or is generated from the first monocular video frame (right). The figure shows the dynamic objects rendered at multiple timesteps from a fixed novel view. The colored lines visualize the 3D point trajectories that emerge from our dynamic model.
  }
\end{centering}


\begin{abstract}
 Dynamic scene reconstruction from casual videos has seen recent remarkable progress. Numerous approaches have attempted to overcome the ill-posedness of the task by distilling priors from 2D foundational models and by imposing hand-crafted regularization on the optimized motion. However, these methods struggle to reconstruct scenes from extreme novel viewpoints, especially when highly articulated motions are present. In this paper, we present \emph{\ourmethod{}} -- a novel approach that leverages a static pre-scan of the dynamic object as an explicit geometric and appearance prior.
 While existing state-of-the-art methods fail to fully exploit the pre-scan,
 \ourmethod{}~leverages our novel setup
 to effectively constrain the solution space and ensure geometrical consistency throughout the sequence.
 The core of our novelty is twofold: first, we establish a grid-structured and surface-aligned model by organizing Gaussian primitives into pixel grids anchored to the object surface.
 Second, by leveraging the grid structure of our primitives, we parameterize motion using a CNN conditioned on those grids, injecting strong implicit regularization and correlating the motion of nearby points. Extensive experiments demonstrate that our method significantly outperforms the current state of the art in rendering quality and 3D tracking accuracy.

\end{abstract}

\section{Introduction}
Reconstruction of dynamic 3D subjects 
from casual video is transformative for many industries, from unlocking new forms of 3D content to metric-space understanding of how objects move, with applications in robotics, augmented reality, and scene analysis.
While significant progress has been made, existing approaches often struggle to maintain high-fidelity reconstruction and consistent novel view synthesis in casual video capture settings, particularly when rendered from extreme viewpoints far from the input camera trajectory.


%
Existing methods for dynamic reconstruction span a spectrum from fully multi-view setups to purely monocular approaches. Multi-view methods~\cite{luiten2023dynamic,joo2015panoptic, wu20234d, Li_STG_2024_CVPR} can achieve excellent results but require expensive synchronized camera arrays that are impractical for casual capture. Monocular methods~\cite{gao2022monocular,park2021nerfies, som2024, lei2024mosca} are highly practical but face severe ill-posedness: infinitely many 3D motions can explain the same 2D observations. In this paper, we propose a novel setting that bridges this gap: monocular dynamic capture combined with a static pre-scan of the object.
The pre-scan is either provided as part of the monocular sequence or generated by an image-to-3D model—a field that continues to advance rapidly. This setting is nearly as practical as monocular, while providing crucial geometric constraints that dramatically reduce ambiguity.

In this paper, we present such an approach by building upon 3D Gaussian Splatting~\cite{kerbl20233d} (3DGS), which has recently emerged as a powerful representation for novel view synthesis. 3DGS represents scenes as collections of colored 3D Gaussians that are rendered via differentiable splatting, enabling real-time rendering and efficient optimization. Recent works have extended 3DGS to dynamic scenes~\cite{luiten2023dynamic,wu20234d,yang2023deformable}, but these typically require multi-view input or struggle with geometric accuracy from novel viewpoints.
We extend this paradigm to the novel pre-scan-plus-monocular setting.
Our approach consists of two main components: (i) constructing structured, surface-aligned Gaussians from the pre-scan by organizing them on pixel grids anchored to the object surface, and (ii) modeling the deformation of Gaussians through a convolutional network (CNN).


A key insight of our approach is that by organizing canonical Gaussians into structured, surface-aligned pixel grids and by supervising them with lifted 2D tracks, we ensure that each primitive persistently represents the same surface point throughout the sequence.
This contrasts with prior 3DGS representations, which typically consist of unordered primitives and/or lack surface representation. While works like PixelSplat~\cite{charatan23pixelsplat} organize Gaussians on pixel grids derived from input images, they do not anchor these primitives to a consistent object manifold. Conversely, while methods such as SuGaR~\cite{guedon2023sugar} and DM4D~\cite{li2024dreammesh4d} encourage Gaussians to align with the object surface, their underlying representation remains unorganized.
Our persistent, surface-aligned, and grid-structured modeling is critical for achieving geometrically faithful reconstructions and high-quality novel view synthesis, particularly from viewpoints not observed in the input video.

Our grid-structured Gaussians are constructed on virtual camera planes oriented toward the canonical pre-scan 3DGS representation. By estimating appearance and depth from these virtual viewpoints, we back-project each pixel to a Gaussian, yielding a set of surface-aligned primitives organized on pixel grids.


Building on this structured representation, we introduce a \motiondip~(\motiondipshort), a CNN that maps the canonical Gaussian grids to per-timestep 6-DOF deformations. By operating on structured pixel grids rather than unordered point sets, our approach leverages the strong spatial inductive bias of Convolutional networks: 
their translation equivariance and local connectivity naturally enforce local smoothness, correlating the motion of adjacent surface points and favoring spatially coherent motion fields, without heavily relying on explicit handcrafted regularization.
Consequently, as demonstrated in Fig.~\ref{fig:teaser}, our design enables geometrically consistent dynamic reconstructions of highly articulated scenes, achieving high-quality synthesis even from extreme novel views.

We evaluate \ourmethod{} on real-world and synthetic captures, and demonstrate that it significantly outperforms prior state-of-the-art monocular dynamic reconstruction methods, including optimization and generative-based approaches, in photometric, perceptual, semantic consistency, and long-range 3D tracking.

%
To summarize, our core contributions are: (\textbf{1}) We introduce a new task of dynamic 3D reconstruction from casual videos with pre-scanned objects~--~a practical setting with significantly reduced ill-posedness. (\textbf{2}) We present a novel canonical model using surface-aligned 3D Gaussians organized into pixel grids, ensuring persistent representation of each surface point. (\textbf{3}) We propose a CNN-based motion parameterization (\motiondipshort) that leverages the convolutional inductive bias for implicit regularization.
(\textbf{4}) On real-world and synthetic benchmarks, we significantly outperform prior state-of-the-art methods by over 1~dB in PSNR with substantial improvements in perceptual (LPIPS), semantic (CLIP), and 3D tracking metrics; systematic ablations confirm that each core design component contributes meaningfully, with removing any one of them causing a noticeable performance drop.

\section{Related Work}
\label{sec:related_work}

\boldparagraph{Dynamic Novel-View Synthesis.}
The field of novel-view synthesis has seen remarkable progress over the years \cite{Lombardi2019NeuralV}, especially after the introduction of Neural Radiance Fields (NeRF)~\cite{mildenhall2020nerf, dynerf, rerf, lin2022enerf, attal2023hyperreel}.
Previous works fall into several categories: (\textbf{a}) Per-timestep methods~\cite{xian2020space} that fit independent representations for each frame without modeling temporal correspondence. (\textbf{b}) Eulerian approaches~\cite{fridovich2023k,cao2023hexplane} that represent scenes on 4D space-time grids, enabling smooth interpolation but lacking explicit correspondence. (\textbf{c}) Canonical-plus-deformation methods~\cite{park2021nerfies,pumarola2021d,park2021hypernerf} that learn a canonical representation and per-frame deformation fields. (\textbf{d}) Template-guided methods~\cite{li2022tava,weng2022humannerf,isik2023humanrf,yang2022banmo} that leverage human body models or other priors.
Related monocular approaches incorporate motion via scene flow or other temporal constraints to enable view and time synthesis from a single posed video \cite{li2023dynibar, li2021neural, dynerf, Gao-ICCV-DynNeRF, gao2022monocular}.
Most related to our work are point-based Lagrangian methods that track individual primitives through time \cite{luiten2023dynamic}.
Our work falls in between the full multi-camera and full monocular settings by taking a static pre-scan and a monocular dynamic sequence as input.
In contrast to these approaches, we initialize canonical, grid-structured, and surface-aligned Gaussians and employ a CNN-based motion prior for regularization.

\boldparagraph{Dynamic 3D Gaussians.}
3D Gaussian Splatting~\cite{kerbl20233d} has emerged as a powerful representation for real-time novel view synthesis. Several works have extended this to dynamic scenes: Dynamic 3D Gaussians~\cite{luiten2023dynamic} tracks persistent Gaussians through time but requires multi-view input; 4D Gaussian Splatting~\cite{wu20234d} and Deformable 3D Gaussians~\cite{yang2023deformable} learn temporal deformations but struggle with geometric consistency from extreme viewpoints; SC-GS~\cite{huang2023sc} introduces sparse control points for editable dynamic scenes. More recently, Shape-of-Motion \cite{som2024} and MoSca \cite{lei2024mosca} achieved SOTA results by leveraging a low-rank motion basis and incorporating foundation model priors for depth and 2D tracking. Their successors, HiMoR \cite{himor} and OriGS \cite{origs}, introduced improvements by hierarchical and orientation-aware motion modeling. Similarly, we incorporate priors from 2D tracking and depth foundation models. Unlike these methods, we leverage a pre-scan prior and surface-aligned pixel-grid organization to ensure geometric consistency even from viewpoints far from the input camera.
We show that SOTA dynamic 3DGS methods fail to effectively leverage the pre-scan information provided within the monocular sequence as an initial static scan preceding the dynamic motion.
Moreover, instead of heavily relying on handcrafted motion priors, we leverage the strong implicit regularization of CNNs by parameterizing motion with a U-Net \cite{ronneberger2015u}.

\boldparagraph{Deep Image Prior.}
\cite{ulyanov2018deep} demonstrated that the architecture of CNNs provides a strong implicit prior for image restoration, even without any training data. This deep prior has been extended to various domains including image denoising~\cite{heckel2019deep}, Bayesian inference~\cite{cheng2019bayesian}, 3D reconstruction~\cite{williams2019deep, mihajlovic2024SplatFields}, and point tracking \cite{dinotracker}. The key insight is that CNN architectures inherently favor smooth, natural signals due to their local connectivity and translation equivariance. We apply this principle to motion estimation: by parameterizing scene motion through a CNN operating on structured canonical parameter grids, we leverage the network's spatial inductive bias to regularize the ill-posed monocular motion estimation problem.

\boldparagraph{Point Tracking.}
Dense point tracking has advanced significantly with learning-based methods. PIPs~\cite{harley2022particle} introduced a transformer-like approach for tracking points through occlusions; TAP-Vid~\cite{doersch2022tap} established benchmarks; TAPIR~\cite{doersch2023tapir}, CoTracker~\cite{karaev2023cotracker}, and AllTracker~\cite{harley2025alltracker} improved accuracy and efficiency. OmniMotion~\cite{wang2023tracking} and DINO-Tracker \cite{dinotracker} are most related to our work, using test-time optimization to estimate dense motion fields, but these methods focus on 2D tracking and require optical flow or video input. In contrast, we leverage 2D tracking predictions as supervision for our 3D motion estimation, and our surface-aligned Gaussians provide 3D correspondence throughout the sequence.







\section{Method}

\begin{figure*}[t!]
  \centering
  \includegraphics[width=1\textwidth]{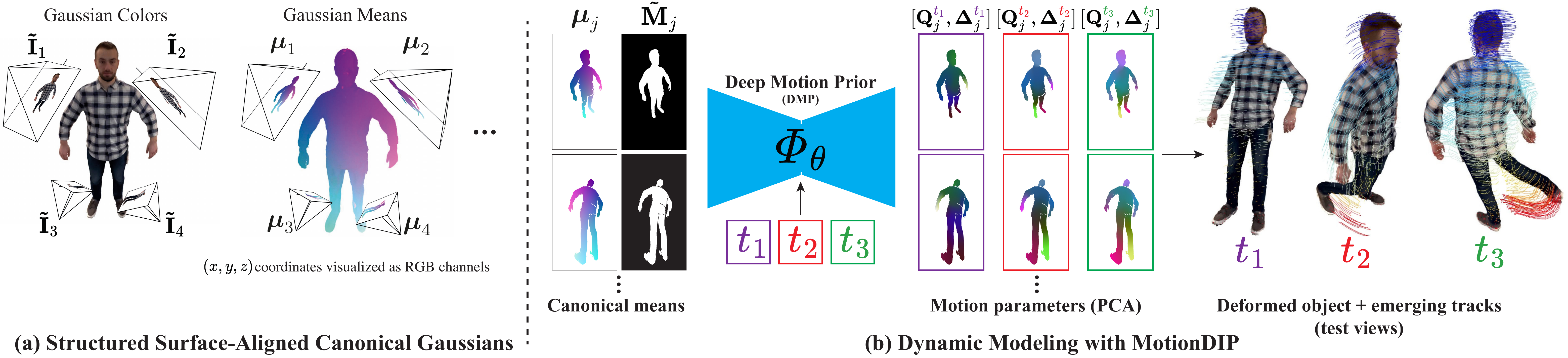}
  \caption{\emph{\ourmethod{} Overview}. (a) We organize our canonical Gaussians into structured pixel grids that reside on virtual cameras surrounding the object, each pixel encoding the parameters of its back-projected 3D Gaussian (\cref{sec:structured-gauss}). (b) To reconstruct the dynamic sequence, we model the object deformation with \motiondip{} $\motiondipmath$ -- a CNN that maps canonical positions $\mugrid_j$ and timestep encodings $\posenc(\timestep)$ to 6-DOF motion parameters $[\rotquat_j, \translation_j]$ (see \cref{sec:dmp}). Timesteps ${\color{t1color}t_1}, {\color{t2color}t_2}, {\color{t3color}t_3}$ are color-coded in the figure. }
  \label{fig:pipeline}
\end{figure*}

Given a monocular video of $T$ frames $\video \coloneqq \{ (\mathframe_{\mathtt{V}}^t, \depth_{\mathtt{V}}^t, \cam_{\mathtt{V}}^t) \}_{t=1}^T$ of a dynamic object, where each timestep includes an RGB frame $\mathframe_{\mathtt{V}}^t$, an estimated depth map $\depth_{\mathtt{V}}^t$, and an estimated camera pose $\cam_{\mathtt{V}}^t$, together with a pre-scan of $R$ frames $\prescan \coloneqq \{ (\prescanframes, \depth_{\mathtt{S}}^r, \cam_{\mathtt{S}}^r) \}_{r=1}^R$ of the same object in a static state (either captured by the monocular camera or generated by an image-to-3D model), our goal is to obtain a complete dynamic 3D reconstruction enabling $360^\circ$-orbiting novel view synthesis. Depth maps and camera poses are obtained from off-the-shelf methods (see \cref{sec:impl}). As illustrated in \cref{fig:pipeline}, our method comprises two core components: (1) constructing canonical, surface-aligned Gaussians organized on pixel grids, where each primitive persistently represents a fixed surface point, and (2) modeling scene dynamics via a \motiondip~(\motiondipshort)---a CNN that predicts per-timestep 6-DOF deformations for each Gaussian.

\subsection{Pre-scan Canonical Model}
\label{sec:method-canonical}

Our canonical model leverages the static pre-scan to construct a structured 3D Gaussian representation that serves as the geometric foundation for dynamic reconstruction. We first build surface-aligned Gaussians organized into pixel grids, then refine them using photometric supervision.
The pre-scan is either captured as part of the monocular sequence or, for in-the-wild fully monocular videos, generated from the first frame using an image-to-3D model~\cite{xiang2025trellis2}. Since image-to-3D generations are not inherently aligned with the video camera, we perform a pose alignment step using MASt3R~\cite{mast3r} correspondences with Perspective-n-Points (PnP) to register the generated mesh to the first frame (see Appendix \ref{sec:supp_alignment} for details).

\subsubsection{Structured Surface-Aligned Canonical Gaussians}

\label{sec:structured-gauss}

We use 3D Gaussian Splatting~\cite{kerbl20233d} to represent the canonical scene. Given the pre-scan $\prescan$, we compute segmentation masks $\masks \coloneqq \{\prescanmasks\}_{r=1}^R$ using SAM2~\cite{ravi2024sam2}. We then fit a 3DGS representation $\gaussinit \coloneqq \{ g_i \coloneqq (\mu_i, s_i, q_i, \alpha_i, c_i, m_i) \}_{i=1}^N$ with $N$ Gaussians. Each Gaussian $g_i$ is parameterized by its mean $\mu_i \in \mathbb{R}^3$, scale $s_i \in \mathbb{R}_+^3$, rotation quaternion $q_i \in \mathbb{R}^4$, opacity $\alpha_i \in [0,1]$, and spherical harmonic coefficients $c_i$ for view-dependent color. Additionally, $m_i \in [0,1]$ is a foreground (FG) probability distilled from $\masks$, used to separate dynamic from static Gaussians. We employ existing techniques for mask distillation \cite{seidenschwarz2025dynomo, zhou2024feature}, which additionally reconstruct per-Gaussian mask channels during 3DGS optimization.

\begin{figure}[t!]
    \centering
    \includegraphics[width=1\textwidth]{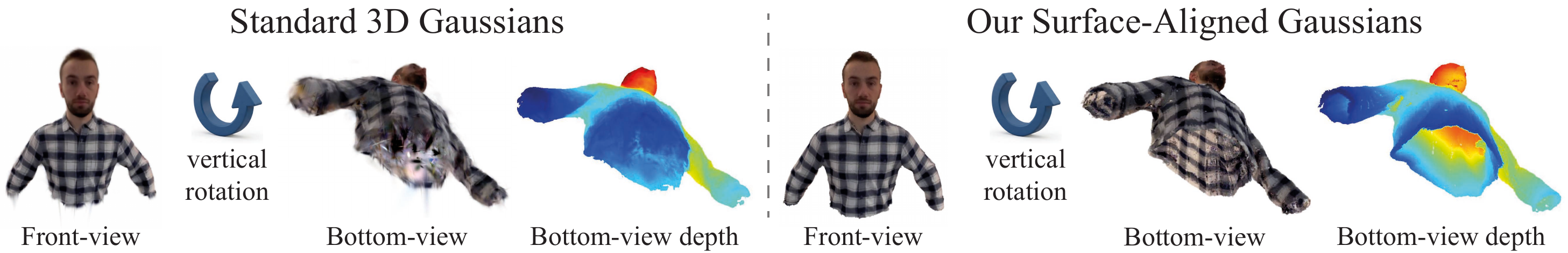}
    \caption{\emph{Surface-aligned Gaussians}. We visualize canonical Gaussians of the upper body. Unlike standard 3DGS (left), where Gaussians are unordered and lack surface representation, our structured Gaussians align with the object's surface, providing a more robust and generalizable representation for dynamic-time reconstruction.
    }
    \label{fig:gauss-surf}
\end{figure}

While standard Gaussian Splatting achieves high-fidelity photometric reconstruction, it lacks an explicit surface representation~\cite{Dai2024GaussianSurfels, Huang2DGS2024} and does not guarantee that primitives persistently correspond to the same surface points over time. To address this, as illustrated in \cref{fig:pipeline}(a) and \cref{fig:gauss-surf}, we organize Gaussians into structured, surface-aligned pixel grids that anchor each primitive to a consistent surface location. We show that this persistent modeling is crucial for faithful and consistent dynamic reconstruction.

To obtain the surface-aligned representation from $\gaussinit$, we first compute a 3D bounding box $\bbox$ around the FG Gaussians $\gaussfg = \{ g_i \in \gaussinit : m_i > 0.5 \}$. For each face $j$ of $\bbox$, we define a virtual stereo camera pair looking inward toward the object center and render $\gaussfg$ from both views. The resulting stereo image pairs are processed by FoundationStereo~\cite{foundationstereo, wolf2024gsmesh} to estimate dense depth maps. We denote the rendered image, mask, and estimated depth from the left view as $\tilde{\mathframe}_{j}$, $\tilde{\mathmask}_{j}$, and $\tilde{\depth}_{j}$, respectively, and extract DINOv2 features $\tilde{\mathbf{F}}_{j}$~\cite{oquab2023dinov2} from $\tilde{\mathframe}_{j}$.
These depth maps define the object surface from each viewpoint. For each face $j$ and each FG pixel $(x, y)$ in $\tilde{\mathframe}_{j}$, we create an isotropic Gaussian $g_{j,x,y}$ with color $c_{j,x,y} \coloneqq \tilde{\mathframe}_{j}[x, y]$, DINOv2 feature $f_{j,x,y} \coloneqq \tilde{\mathbf{F}}_{j}[x, y]$, identity rotation $q_{j,x,y} \coloneqq [1, 0, 0, 0]$, fixed opacity $\alpha_{j,x,y} \coloneqq 0.98$, and position/scale derived from the estimated depth $\tilde{\depth}_{j}[x, y]$:
\begin{align}
    \mu_{j, x, y} \coloneqq \tilde{\cam}_{j}^{-1}(x, y)\tilde{\depth}_{j} [x, y]\,, \qquad s_{j, x, y} \coloneqq \frac{\tilde{\depth}_{j} [ x, y ] }{ \focal } \cdot 0.95\,,
\end{align}
where $\tilde{\cam}_{j}^{-1}(x, y) \in \mathbb{R}^3$ denotes pixel $(x,y)$ unprojected to unit depth (i.e., $z=1$) in camera space, and $\focal$ is the focal length. We denote the resulting set of surface-aligned Gaussians as $\gausscanon \coloneqq \{ g_{j,x,y} \}$.


\subsubsection{Canonical Refinement}

The depth estimates from FoundationStereo provide a good initialization for $\gausscanon$, but lack high-frequency surface details (see Appendix \ref{sec:supp_depth_ref} for visualizations). To recover fine-grained geometry, we refine the depth maps $\tilde{\depth}_{j}$ and colors $\tilde{\mathframe}_{j}$ that parameterize the canonical Gaussians by minimizing a photometric loss on the pre-scan views $\prescan$. For each pre-scan view $r$, we define:
\begin{equation}
    \losscanonref^r \coloneqq \losslone(\hat{\mathframe}_{\mathtt{S}}^r, \mathframe_{\mathtt{S}}^r) + \lambdassim \lossssim(\hat{\mathframe}_{\mathtt{S}}^r, \mathframe_{\mathtt{S}}^r) + \lambdatv \sum_{j} \losstv(\tilde{\mathframe}_{j}) \,,
    \label{eq:canonref}
\end{equation}
where $\mathframe_{\mathtt{S}}^r$, $\hat{\mathframe}_{\mathtt{S}}^r$ are the ground-truth and rendered images from pre-scan camera $\cam_{\mathtt{S}}^r$, respectively. $\losstv$ is a total variation regularizer that encourages spatial smoothness in the parameter grids.
$\lambda_{*}$ are trade-off parameters (values in Appendix \ref{sec:supp_hyperparams}).
The full canonical refinement loss averages over all pre-scan views: $\losscanonref \coloneqq \frac{1}{R} \sum_{r=1}^{R} \losscanonref^r$.

\subsection{ \motiondip }

\label{sec:dmp}

Estimating 3D motion from a monocular video is inherently ill-posed: many deformations satisfy the photometric loss while significantly distorting geometry. Existing methods address this with handcrafted priors such as as-rigid-as-possible regularization~\cite{huang2023sc} or low-rank motion~\cite{som2024}. Instead, we harness the deep implicit prior of CNNs by leveraging the pixel-grid organization of $\gausscanon$.

We design a \motiondip~(\motiondipshort) $\motiondipmath$ that takes as input the canonical position grids $\mugrid_j \in \mathbb{R}^{H \times W \times 3}$, where $\mugrid_j[x,y] \coloneqq \mu_{j,x,y}$, and outputs per-pixel 6-DOF deformations. For timestep $t$:
\begin{align}
    (\rotquat^t_j, \translation_j^t) \coloneqq \motiondipmath(\mugrid_j, t) \,,
\end{align}
where $\rotquat_j^t \in \mathbb{R}^{H \times W \times 4}$ and $\translation_j^t \in \mathbb{R}^{H \times W \times 3}$ are per-Gaussian rotation quaternions and translations, respectively. The deformed positions are computed as:
\begin{align}
    \mugrid^t_j \coloneqq \text{rot}(\mugrid_{j}, \rotquat_j^t) + \translation_j^t \,,
\end{align}
where $\text{rot}(\cdot, \cdot)$ applies the quaternion rotation pixel-wise. To prevent content duplication across parameter grids, we apply de-duplication masks on output grids $\mugrid^t_j$ (see Appendix \ref{sec:supp_dedup} for details). The network $\motiondipmath$ is initialized to predict identity transformations, ensuring optimization starts from the canonical pose. The CNN's spatial inductive bias naturally enforces smooth, locally coherent deformations.


\subsection{Optimization}

This section enumerates the loss terms used to optimize our \motiondip.

\boldparagraph{Photometric Loss.} For each timestep $t$, we render the deformed Gaussians from the input camera $\cam_{\mathtt{V}}^t$, producing an image $\hat{\mathframe}_{\mathtt{V}}^t$. We minimize the difference between the rendered and observed images using the standard 3DGS loss~\cite{kerbl20233d}:
\begin{align}
    \lossphoto^t \coloneqq \losslone(\hat{\mathframe}_{\mathtt{V}}^t, \mathframe_{\mathtt{V}}^t) + \lambdassim \lossssim(\hat{\mathframe}_{\mathtt{V}}^t, \mathframe_{\mathtt{V}}^t) \,,
    \label{eq:lphoto}
\end{align}
where $\mathframe_{\mathtt{V}}^t$ is the ground-truth frame from $\video$. 

\boldparagraph{Point Tracking Loss.} We leverage the prior of a pre-trained 2D point tracker (AllTracker~\cite{harley2025alltracker}) to supervise the 2D foreground motion predicted by \motiondipshort. Given $K$ foreground query points $\{\point_k^1\}_{k=1}^K$ sampled at the reference frame ($t=1$), AllTracker provides ground-truth 2D correspondences $\point_k^t$ for each timestep $t$. Our model's prediction $\hat{\point}_k^t$ is obtained by rasterizing the 2D positions of dynamic Gaussians at time $t$ onto the reference frame \cite{som2024}. The loss is defined as:
\begin{align}
    \losstrack^t \coloneqq \frac{1}{K} \sum_{k=1}^K \| \point_k^t - \hat{\point}_k^t \|_1 \, .
\end{align}

\boldparagraph{Depth Loss.} For each timestep $t$, we render the depth of the deformed Gaussians from camera $\cam_{\mathtt{V}}^t$, producing a depth map $\hat{\depth}_{\mathtt{V}}^t$. Let $\mathmask_{\mathtt{FG}}^t$ denote the intersection of the ground-truth and rendered foreground masks. We supervise the rendered depth against the input depth $\depth_{\mathtt{V}}^t$ within this region:
\begin{align}
    \lossdepth^t \coloneqq \frac{1}{\|\mathmask_{\mathtt{FG}}^t\|_1} \| (\hat{\depth}_{\mathtt{V}}^t - \depth_{\mathtt{V}}^t) \odot \mathmask_{\mathtt{FG}}^t \|_1 \, ,
\end{align}
where $\odot$ denotes element-wise multiplication.

\boldparagraph{Depth Reprojection Loss.} 
We additionally supervise the \emph{reprojected} depth of dynamic points. Given a foreground query point $\point_k^1$ in the reference frame ($t=1$) and its tracked correspondence $\point_k^t$ in frame $t$, we obtain the ground-truth depth by sampling the input depth map: $\depth_{\mathtt{V}}^t[\point_k^t]$.
To obtain our model's prediction, we render for each Gaussian its depth at time $t$ but at the 2D location it occupies in frame $1$, yielding the reprojected depth map $\hat{\depth}^{t \rightarrow 1}$. The loss encourages consistency between tracked correspondences and predicted motion:
\begin{align}
    \lossreproj^t \coloneqq \frac{1}{K} \sum_{k=1}^{K} \| \depth_{\mathtt{V}}^t[\point_k^t] - \hat{\depth}^{t \rightarrow 1} [ \point_k^1 ] \|_1 \, .
\end{align}

\boldparagraph{Isometry Losses.} To preserve geometric consistency during deformation, we introduce multi-scale isometry losses that encourage distance preservation between Gaussian pairs at different spatial scales. For notational clarity, we re-index the canonical Gaussians $g_{j,x,y} \in \gausscanon$ with a single index $\mathfrak{i}$, writing $g_\mathfrak{i}$ to denote an arbitrary Gaussian, $\mu_\mathfrak{i}$ for its canonical position, and $\mu_\mathfrak{i}^t$ for its deformed position at timestep $t$ as predicted by \motiondipshort.

\emph{Coarse isometry.} This loss prevents global geometry drift and maintains the overall object shape. We randomly sample 1\% of all Gaussians from $\gausscanon$, forming a sparse subset $\pcoarse$. For each Gaussian $g_\mathfrak{i} \in \pcoarse$, we enforce that distances to its nearest neighbors $\nncoarse(g_\mathfrak{i})$ in $\pcoarse$ are preserved after deformation~\cite{luiten2023dynamic}:
\begin{align}
    \losscoarseiso^t
    \coloneqq \sum_{\substack{g_\mathfrak{i} \in \pcoarse \\ g_\mathfrak{j} \in \nncoarse(g_\mathfrak{i})}}
    \frac{w_{\mathfrak{ij}}}{|\pcoarse|} \left| \| \mu_\mathfrak{i} - \mu_\mathfrak{j} \|_2 - \| \mu_\mathfrak{i}^t - \mu_\mathfrak{j}^t \|_2 \right| \, ,
\end{align}
where $w_{\mathfrak{ij}} \coloneqq \mathcossim(f_\mathfrak{i}, f_\mathfrak{j})$ weights the constraint by the DINO feature similarity between Gaussians $g_\mathfrak{i}$ and $g_\mathfrak{j}$, allowing semantically similar and nearby regions to deform together.

\emph{Dense isometry.} This loss enforces local surface rigidity across all canonical Gaussians. For each Gaussian $g_\mathfrak{i} \in \gausscanon$, we penalize distance changes to its immediate neighbors $\nn(g_\mathfrak{i})$:
\begin{align}
    \lossdenseiso^t
    \coloneqq \sum_{\substack{g_\mathfrak{i} \in \gausscanon \\ g_\mathfrak{j} \in \nn(g_\mathfrak{i})}}
    \frac{w'_{\mathfrak{ij}}}{|\gausscanon|} \left| \| \mu_\mathfrak{i} - \mu_\mathfrak{j} \|_2 - \| \mu_\mathfrak{i}^t - \mu_\mathfrak{j}^t \|_2 \right| \, ,
\end{align}
where $w'_{\mathfrak{ij}} \coloneqq \exp(-\beta \| \mu_\mathfrak{i} - \mu_\mathfrak{j} \|_2^2)$ assigns higher weights to closer neighbors, with $\beta$ controlling the spatial falloff (see Appendix \ref{sec:supp_hyperparams} for the value).

\boldparagraph{Rigidity Loss.} To further regularize the deformation field, we encourage \motiondipshort~to predict similar rotations for spatially close Gaussians. For each Gaussian $g_\mathfrak{i} \in \pcoarse$, we penalize rotation differences with its neighbors:
\begin{align}
    \lossrigid^t \coloneqq \sum_{\substack{g_\mathfrak{i} \in \pcoarse \\ g_\mathfrak{j} \in \nncoarse(g_\mathfrak{i})}} \frac{w_{\mathfrak{ij}}}{|\pcoarse|} \| q_\mathfrak{i}^t - q_\mathfrak{j}^t \|_1 \, .
\end{align}

\boldparagraph{Final Objective.} For each timestep $t$, we define the total loss:
\begin{multline}
    \mathcal{L}^t \coloneqq \lossphoto^t + \lambdatrack \losstrack^t + \lambdadepth \lossdepth^t + \lambdareproj \lossreproj^t \\
    + \lambdacoarseiso \losscoarseiso^t + \lambdadenseiso \lossdenseiso^t + \lambdarigid \lossrigid^t \, ,
    \label{eq:lfinal}
\end{multline}
where $\lambda_*$ denote the relative weights between loss terms. We use a fixed set of $\lambda_*$ across all experiments (see Appendix \ref{sec:supp_hyperparams} for details). Our full objective averages over all timesteps: $\mathcal{L} \coloneqq \frac{1}{T} \sum_{t=1}^{T} \mathcal{L}^t$.

\begin{figure*}[t!]
  \centering
  \includegraphics[width=1\textwidth]{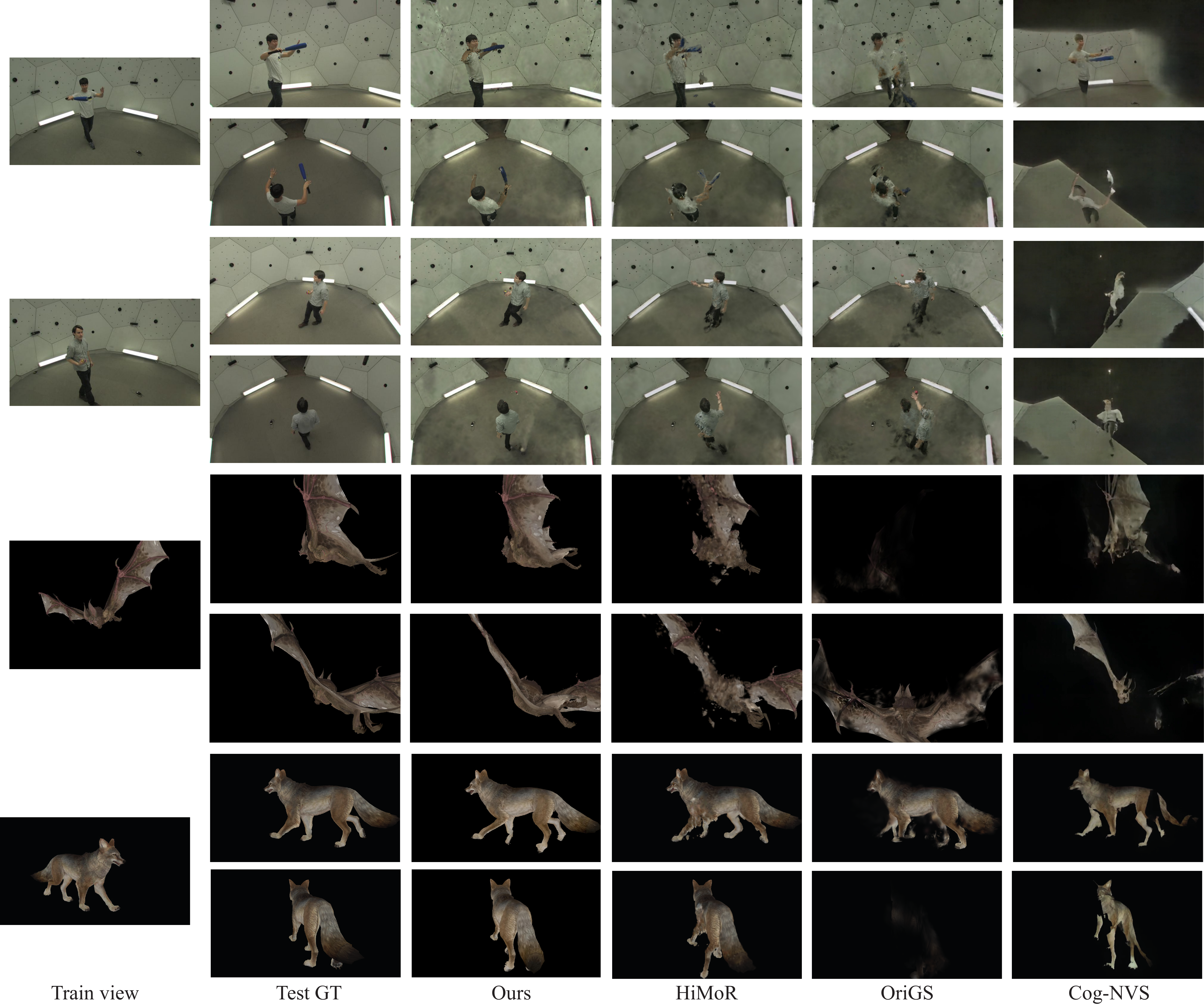}
  \caption{\emph{Qualitative results}. The first column depicts the training view from the monocular sequence; the second column depicts the ground-truth testing view at the same timestep. Novel views are selected at extreme angles to evaluate the completeness of dynamic 3D reconstructions. Our method drastically outperforms the baselines in maintaining a consistent object geometry and sharp appearance, while accurately modeling the scene dynamics. See our website for video results.
  }
  \label{fig:qual}
\end{figure*}

\begin{table*}[t!]
\centering
\caption{\emph{Quantitative evaluation}. We evaluate our method on real-world (Panoptic Studio) and synthetic (Truebones) benchmarks against SOTA monocular dynamic reconstruction methods, as discussed in \cref{sec:results}. All competitors are adapted to the pre-scan-plus-monocular setting and are conditioned on the same preprocessed inputs of 2D tracks, depth, and camera estimates. Our method achieves SOTA performance over optimization and generative-based competitors in all metrics: image reconstruction quality (PSNR, SSIM), perceptual quality (LPIPS), and semantic consistency (CLIP).
}
\label{tab:quant}
\begin{adjustbox}{width=\textwidth}
\begin{tabular}{lcccc cccc}
\toprule
& \multicolumn{4}{c}{\textbf{Panoptic Studio }} & \multicolumn{4}{c}{\textbf{Truebones}} \\
\cmidrule(lr){2-5} \cmidrule(lr){6-9}
\textbf{Method} & \textbf{mPSNR} $\uparrow$ & \textbf{mSSIM} $\uparrow$ & \textbf{mLPIPS} $\downarrow$ & \textbf{CLIP} $\uparrow$ & \textbf{mPSNR} $\uparrow$ & \textbf{mSSIM} $\uparrow$ & \textbf{mLPIPS} $\downarrow$ & \textbf{CLIP} $\uparrow$ \\
\midrule

HiMoR \cite{himor}   & 18.403 & 0.415 & 0.252 & 0.902 & 21.544 & 0.496 & 0.435 & 0.845 \\
OriGS \cite{origs}  & 17.707     & 0.418    & 0.289    & 0.888    & 20.693     & 0.486    & 0.512    & 0.796    \\
Cog-NVS \cite{cog-nvs} & 16.904 & 0.401 & 0.329 & 0.876 & 21.062 & 0.495 & 0.486 & 0.811 \\

\textbf{Ours} & \textbf{19.414} & \textbf{0.447} & \textbf{0.220} & \textbf{0.929} & \textbf{21.756} & \textbf{0.505} & \textbf{0.378} & \textbf{0.876}

\\
\bottomrule
\end{tabular}
\end{adjustbox}
\end{table*}

\section{Experiments}

\subsection{Implementation details}
\label{sec:impl}

\PAR{Datasets.}
We evaluate our method on two datasets: the Panoptic Studio dataset~\cite{joo2015panoptic} and a synthetic dynamic animals dataset that we generate using Truebones \cite{truebones2025}. For Panoptic Studio, we use sequences from the sports subset containing diverse single-subject human motions, including juggling, softball, tennis and boxes. Each sequence contains 150 frames at 30 FPS captured by 31 synchronized HD cameras.
We simulate the monocular setting by using a single selected input view for training and evaluating on the remaining 30 held-out cameras. For the pre-scan, we use all the cameras from the first timestep. For Truebones, since the data is synthetic, we render images and ground-truth depths from known camera parameters for 7 synthetic animated characters for the pre-scan, the dynamic monocular sequence, and the testing views, allowing us to validate our approach under controlled conditions. For 3D tracking evaluation on Panoptic Studio, we use annotations from TAP-Vid-3D \cite{koppula2024tapvid3d}. For Truebones, we extract the ground-truth tracks from dynamic mesh vertices. See Appendix \ref{sec:supp_tracking_eval} for more evaluation and benchmarks details.

\PAR{Preprocessing.}
We obtain FG masks using SAM2~\cite{ravi2024sam2}; dense 2D point tracking supervision with AllTracker \cite{harley2025alltracker}.
For casual videos, depth and cameras are estimated using ViPE~\cite{huang2025vipe}. For TrueBones, we use the GT cameras and rendered GT depth. For Panoptic Studio, we render stereo cameras for the dynamic sequence using the reconstructions of \cite{luiten2023dynamic}
and obtain per-frame depth using~\cite{foundationstereo}.

See Appendix for architecture, hyperparameters and other implementation details.




\subsection{Evaluation metrics}
We evaluate and compare the performance of \ourmethod{} in two primary tasks: (i) novel view synthesis quality, and (ii) long-range 3D tracking accuracy.
For (i), we use standard image quality metrics: PSNR (Peak Signal-to-Noise Ratio) for pixel-level accuracy, SSIM~\cite{wang2004image} for structural similarity, and LPIPS~\cite{zhang2018perceptual} for perceptual quality using deep features. Additionally, we compute CLIP score~\cite{radford2021learning} to measure high-level semantic consistency between rendered and ground-truth views. We measure only the FG reconstruction quality in the first 3 metrics by using ground-truth segmentation masks.
For 3D tracking evaluation, we use metrics from \cite{som2024}, reporting the 3D end-point-error (EPE), and the fraction of points that fall within 5cm and 10cm thresholds of the ground-truth 3D positions, denoted as $\delta_{0.05}$ and $\delta_{0.1}$, respectively.



\subsection{Baselines}
We compare against two categories of methods: (1) Optimization-based approaches including HiMoR~\cite{himor}, which extends \cite{som2024} with hierarchical motion representations, and OriGS \cite{origs}, which utilizes orientation fields and motion scaffolds from \cite{lei2024mosca}, and (2) a generative method Cog-NVS~\cite{cog-nvs}, which uses a "warp-inpaint" approach with video diffusion models for novel view synthesis.
The pre-scan is provided to all baselines within the monocular video as a sub-sequence preceding the dynamic motion. For a fair comparison, all methods are conditioned on the same processed data as ours, including 2D tracks, depths, and camera poses.


\subsection{Results}
\label{sec:results}

We evaluate our method on the Panoptic Studio and Truebones datasets, comparing it against state-of-the-art optimization-based (HiMoR~\cite{himor}, OriGS~\cite{origs}) and generative (Cog-NVS~\cite{cog-nvs}) methods. Importantly, to assess the complete dynamic reconstruction of the methods, we select novel views at extreme angles that differ significantly from the input training views.

\boldparagraph{Quantitative Comparison.} As shown in \cref{tab:quant}, our method significantly outperforms all competitors across all reconstruction metrics on both datasets. On the real-world Panoptic Studio dataset, which presents highly complex and non-rigid deformations, we achieve a substantial improvement over the nearest competitor (HiMoR) in both photometric reconstruction quality (over 1.0 dB in PSNR and about 13\% reduction in LPIPS error) and semantic consistency in CLIP score. The performance gap is larger on the real-world data, highlighting our model's superior ability to handle the complexity and the noise of such captures. In \cref{tab:tracking}, we evaluate the long-range 3D tracking performance of \ourmethod{} compared to SOTA monocular reconstruction methods which output track estimates. As seen, \ourmethod{} outperforms both competitors in all tracking metrics. See our website for reconstruction videos and tracking visualizations.

\boldparagraph{Qualitative Comparison.} Visual comparisons in \cref{fig:qual} further demonstrate \ourmethod's superior performance over the baselines. Our method maintains superior geometric consistency and models complex motions with high fidelity, preserving sharp textures even when rendered from extreme novel views. In contrast, competitors frequently suffer from severe artifacts, such as blurry textures, missing or extremely distorted geometry, and highly inaccurate motion (see our website for videos). Notably, even though we apply test-time fine-tuning on Cog-NVS for each sequence, it fails to plausibly inpaint the scene under the extreme warping required for our target novel viewpoints. Our method succeeds in leveraging the pre-scan as a geometric prior through its persistent, surface-aligned modeling of primitives and strong regularization from \motiondipshort. Crucially, despite providing the pre-scan to the competitors as a part of the monocular video, they fail to effectively exploit this explicit prior.

\begin{figure*}[t!]
  \centering
  \includegraphics[width=1\textwidth]{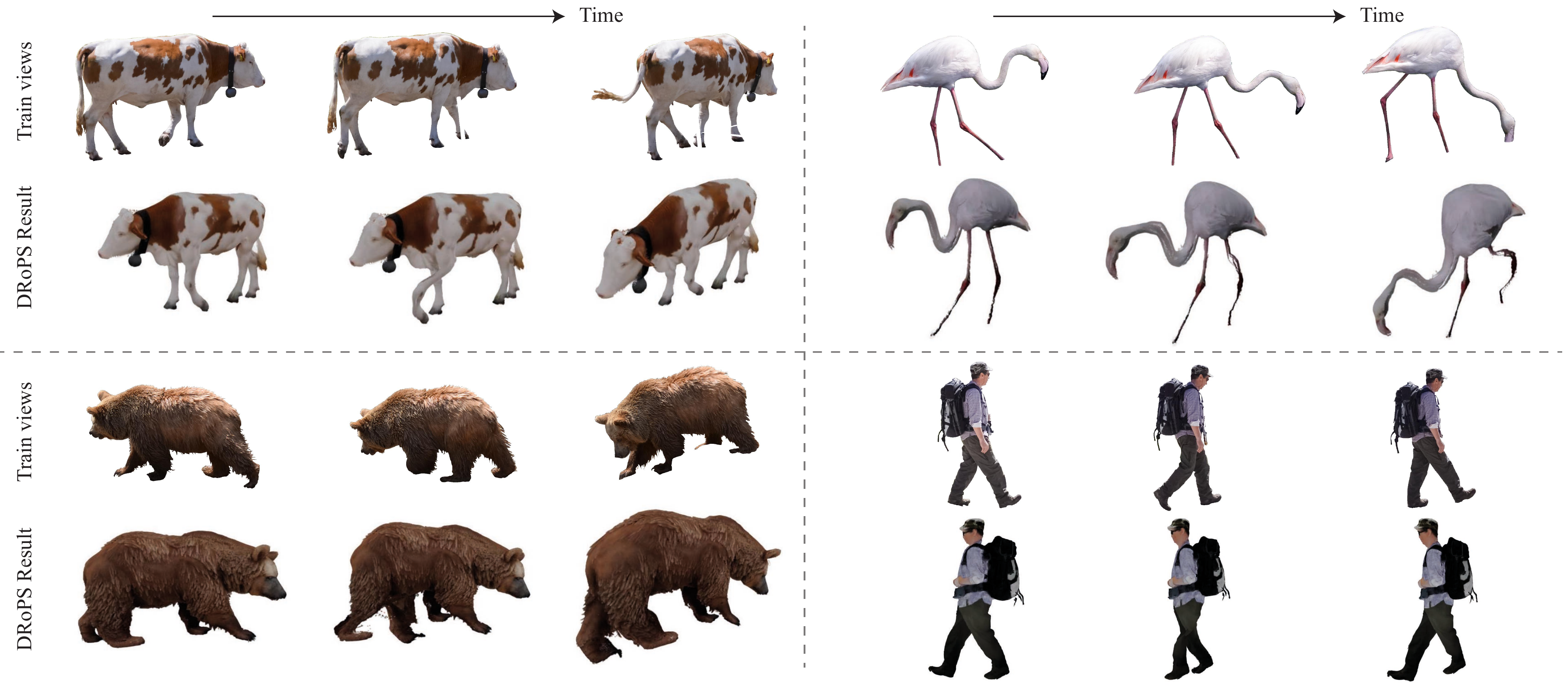}
  \caption{\ourmethod{} achieves high-quality dynamic 3D reconstruction on \emph{in-the-wild monocular videos} by generating the pre-scan with an image-to-3D model \cite{xiang2025trellis2}. Our novel views are rendered from viewpoints that differ significantly from those in the training set. See our website for full video and additional results. }
  \label{fig:gen-app}
\end{figure*}

\begin{figure*}[t!]
  \centering
  \includegraphics[width=1\textwidth]{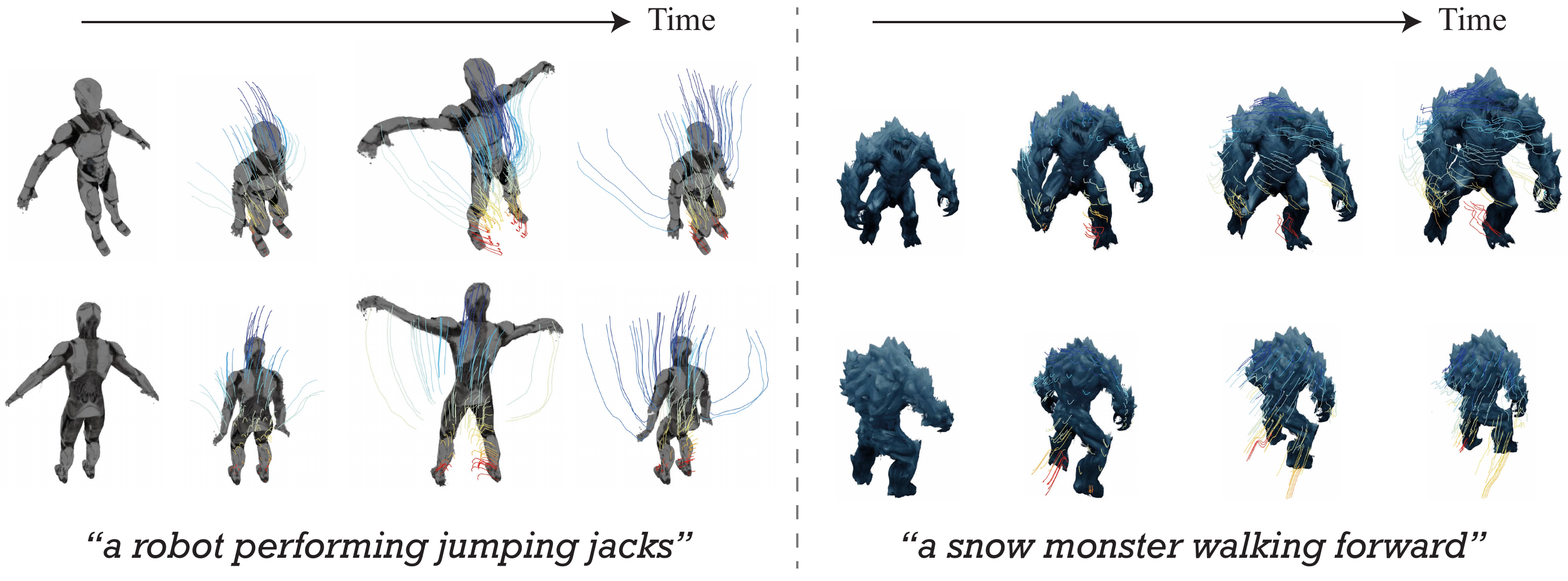}
  \caption{\emph{Text-to-4D application.} Each column and row correspond to a fixed timestep and viewpoint, respectively. The colored lines visualize the 3D trajectories emerging from our representation. See our website for videos.}
  \label{fig:text-2-4d}
\end{figure*}

\subsection{Additional Results}

\boldparagraph{In-The-Wild Videos.} \label{sec:in-the-wild}
Our method can operate on casual, fully monocular videos where the pre-scan is not captured by the camera. To obtain the pre-scan, we apply an image-to-3D model \cite{xiang2025trellis2} on the first frame. As demonstrated in \cref{fig:gen-app}, \ourmethod{} achieves high-quality, complete reconstruction of dynamic monocular videos, capturing the motion and appearance of the objects with high fidelity. See our website for full video visualizations and additional results.

\boldparagraph{Text-to-4D.}
\ourmethod{} can be used for Text-to-4D generation: given a text prompt describing a dynamic subject, we generate a video using an off-the-shelf text-to-video model. We then use the "In-The-Wild Videos" approach for dynamic 3D reconstruction. We demonstrate our Text-to-4D results in \cref{fig:text-2-4d}.

\begin{figure}[b]
    \centering
    \includegraphics[width=0.625\textwidth]{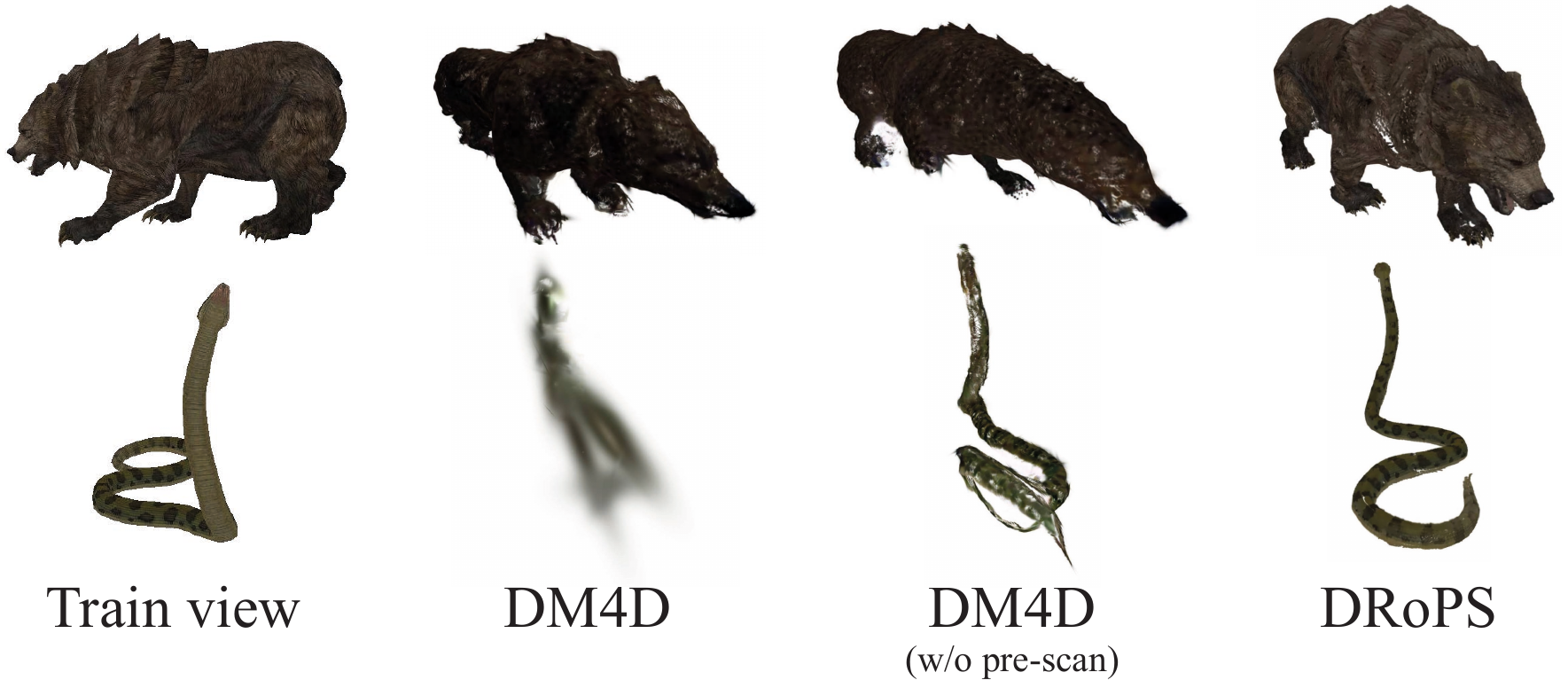}
    \caption{We additionally compare to DM4D \cite{li2024dreammesh4d}, with and without providing them the pre-scan. DM4D results in distorted geometry, inaccurate motion, and visual artifacts.
    In contrast, \ourmethod{} achieves accurate, consistent and realistic reconstructions.}
    \label{fig:dm4d-comp}
\end{figure}

\boldparagraph{Additional Comparison to DreamMesh4D.} In \cref{fig:dm4d-comp}, we provide a qualitative comparison against DM4D \cite{li2024dreammesh4d}.
We evaluate DM4D both with and without a pre-scan. While DM4D suffers from distorted geometry, inaccurate motion estimation, and significant visual artifacts, \ourmethod{} successfully leverages the pre-scan prior to achieve accurate, consistent, and realistic reconstructions.


\subsection{Ablations}

\begin{figure}[H]
  \centering
  \includegraphics[width=0.65\textwidth]{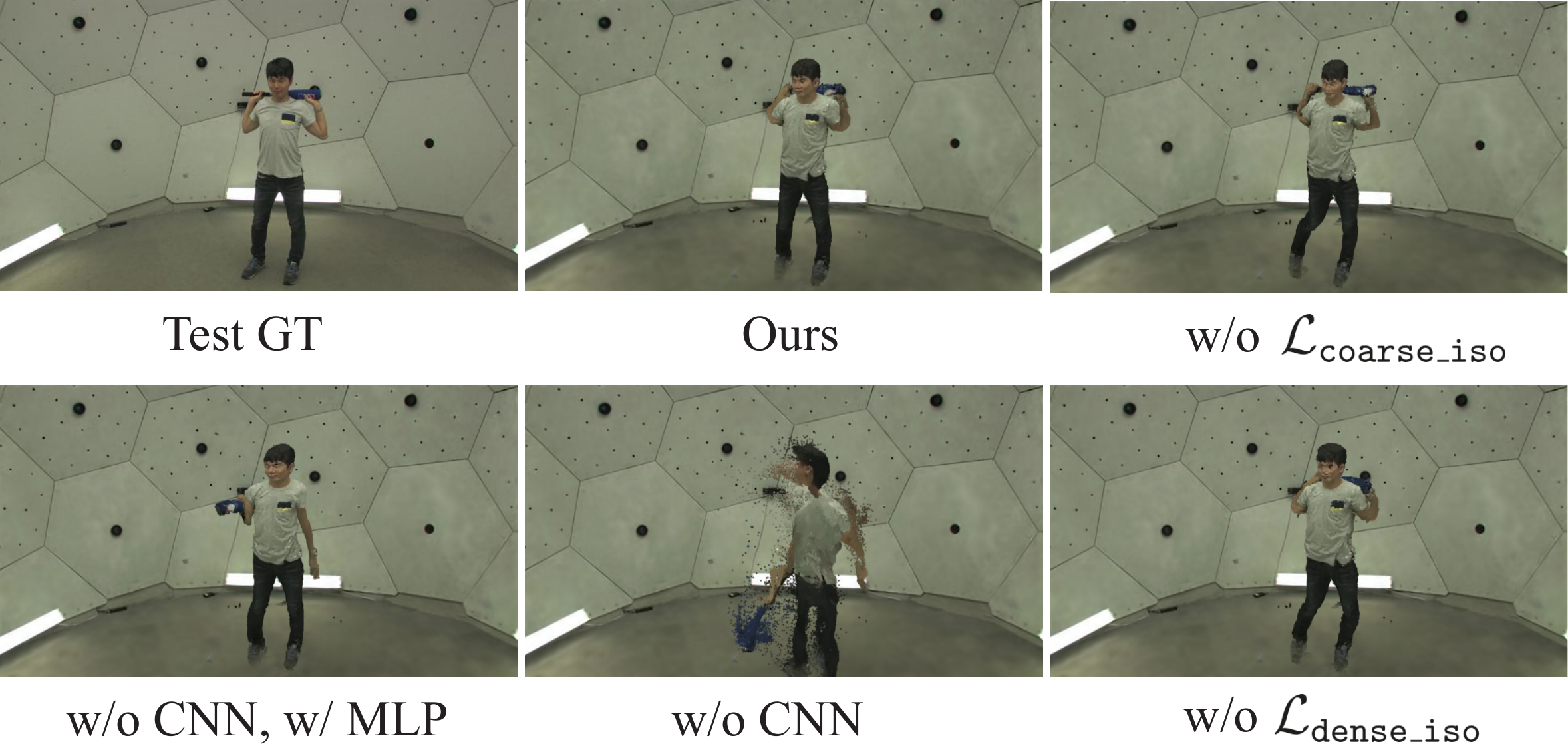}
  \caption{\emph{Ablation Results}. Removing \motiondipshort{} ("w/o CNN") overfits to the training view, failing to capture the motion and geometry. Replacing the CNN with an MLP improves over it, but still suffers from geometry distortion and inaccurate motion. $\losscoarseiso$ is crucial for global geometry preservation, while $\lossdenseiso$ has a relatively small impact since the inductive bias of \motiondipshort~ introduces strong local motion correlation. 
  }
  \label{fig:ablations}
\end{figure}

We ablate the key components of our design on the Panoptic Studio dataset, as demonstrated in \cref{tab:ablations,fig:ablations}. Notably, ablating each component results in a performance drop across all metrics, demonstrating the effectiveness of each component in the overall framework.

The most impactful component of \ourmethod{} is the CNN-based motion parametrization with \motiondip{}. Removing \motiondipshort{} entirely ("w/o \motiondipshort{}") and directly optimizing the per-Gaussian deformation grid drastically drops in performance across all metrics. Replacing the CNN with an MLP ("w/o \motiondipshort{}, w/ MLP") improves over direct optimization. However, its performance remains significantly inferior to our full method. This demonstrates that the spatial inductive bias of CNNs is crucial for regularizing the ill-posed monocular reconstruction problem and for accurately capturing the scene motion.

The coarse isometry loss $\losscoarseiso$ has a substantial impact on geometry preservation, as "w/o $\losscoarseiso$" significantly harms all metrics. Interestingly, the dense isometry loss $\lossdenseiso$ has a relatively small effect. This highlights the strength of CNN's inductive bias to regularize and implicitly correlate local motion.

\begin{table}[t!]
\centering
\caption{Tracking Performance Comparison. We evaluate the tracking accuracy of \ourmethod{} against state-of-the-art monocular reconstruction methods using End-to-Point Error (EPE) and the fraction of points with error below 0.05m ($\delta_{0.05}$) and 0.1m ($\delta_{0.1}$).
Our method consistently outperforms baselines across both datasets.
}
\label{tab:tracking}
\begin{adjustbox}{width=0.55\columnwidth}
\begin{tabular}{l ccc c ccc}
\toprule
& \multicolumn{3}{c}{\textbf{Panoptic Studio}} & & \multicolumn{3}{c}{\textbf{Truebones}} \\
\cmidrule{2-4} \cmidrule{6-8}
\textbf{Method} & \textbf{EPE} $\downarrow$ & $\bm{\delta_{0.05}}$ $\uparrow$ & $\bm{\delta_{0.1}}$ $\uparrow$ & & \textbf{EPE} $\downarrow$ & $\bm{\delta_{0.05}}$ $\uparrow$ & $\bm{\delta_{0.1}}$ $\uparrow$ \\
\midrule
HiMoR \cite{himor}  & 0.139 & 0.335 & 0.564 & & 0.078 & 0.609 & 0.770 \\
OriGS \cite{origs}  & 0.186 & 0.206 & 0.419 & & 0.245 & 0.244 & 0.368 \\
\textbf{Ours}        & \textbf{0.099} & \textbf{0.563} & \textbf{0.743} & & \textbf{0.070} & \textbf{0.691} & \textbf{0.842} \\
\bottomrule
\end{tabular}
\end{adjustbox}
\end{table}

\begin{table}[H]
\centering
\caption{\emph{Ablation Study on Panoptic Studio}. Each component's ablation results in a performance drop, highlighting their contribution. The most impactful design components are \motiondip~ and $\losscoarseiso$.}
\label{tab:ablations}
\begin{adjustbox}{width=0.75\columnwidth}
\begin{tabular}{lccccccc}
\toprule
\textbf{Method} & \textbf{PSNR} $\uparrow$ & \textbf{SSIM} $\uparrow$ & \textbf{LPIPS} $\downarrow$ & \textbf{CLIP} $\uparrow$ & \textbf{EPE} $\downarrow$ & $\bm{\delta_{0.05}}$ $\uparrow$ & $\bm{\delta_{0.1}}$ $\uparrow$ \\
\midrule

w/o \motiondipshort          & 18.289 & 0.409 & 0.313 & 0.9 & 0.274 & 0.131 & 0.267 \\
w/o \motiondipshort, w/ MLP           & 18.853 & 0.435 & 0.236 & 0.919 & 0.172 & 0.231 & 0.433 \\
w/o $\losscoarseiso$       & 18.694 & 0.426 & 0.246 & 0.917 & 0.156 & 0.331 & 0.547 \\
w/o $\lossdenseiso$ & 19.218 & 0.433 & 0.227 & 0.926 & 0.157 & 0.346 & 0.558 \\
w/o $\lossrigid$          & 19.257 & 0.440 & 0.223 & 0.927 & 0.147 & 0.383 & 0.599  \\
\textbf{Ours}         & \textbf{19.414} & \textbf{0.447} & \textbf{0.220} & \textbf{0.929} & \textbf{0.099} & \textbf{0.563} & \textbf{0.743} \\

\bottomrule
\end{tabular}
\end{adjustbox}
\end{table}



\vspace{-0.3cm}
\section{Discussion and Conclusions}
In this work, we have introduced a novel approach for complete dynamic 3D reconstruction from monocular video with a static pre-scan, enabling high-quality novel view synthesis. Our key innovations include: surface-aligned Gaussians organized into pixel grids for persistent surface representation, and a CNN-based motion parameterization that provides strong implicit regularization through its spatial inductive bias.

While \ourmethod{} achieves excellent results, it is not without limitations. First, it currently handles single foreground subjects; extending to multi-object scenes with independent motions would require additional segmentation and per-object motion modeling.
Second, \ourmethod{} cannot handle transparencies or increasing topologies (\eg fire).
Third, our method relies on 2D point trackers and depth estimators for supervision, inheriting any errors from these upstream predictions. We believe these limitations are seeds for exciting future research directions.

We demonstrated the strengths of \ourmethod{} through extensive experiments and evaluations. We showed that \ourmethod{} significantly outperforms prior SOTA reconstruction methods in 3D tracking and novel-view synthesis, including from extreme novel views where geometric consistency is crucial. Our method offers avenues for applications beyond novel view synthesis, such as 3D animation, motion analysis, and content creation from text or casual video capture.


%
%
\bibliographystyle{splncs04}
\bibliography{main}



\appendix
\newpage

\phantomsection
\section{Appendix}
\label{sec:supp_implementation}

\subsection{Architecture.}
\label{sec:supp_arch}
Our \motiondip{} $\motiondipmath$ is implemented as a U-Net style encoder-decoder with skip connections \cite{ronneberger2015u, ulyanov2018deep}, operating on the H×W pixel grids of canonical positions. The network takes the concatenation of canonical position grids $\mugrid_j$, masks grid $\tilde{\mathmask}_j$, and positional-encoded timestep $\posenc(t)$ (with 4 frequencies, spatially broadcasted and concatenated at each $(h, w$)) as input, and outputs 7-channel grids representing quaternion rotations (4 channels) and translations (3 channels). We use 4 encoder blocks with feature dimensions [16, 32, 128, 128], with skip connections of 1x1 convolutions of feature dimension 4. The output of the skip connections are concatenated with the corresponding decoder block's output channels. Each encoder block consists of a 3x3 convolutional layer, followed by a BatchNorm and a LeakyReLU activation. The decoder blocks are symmetrical to the encoder blocks, using nearest neighbor interpolation for upsampling. Crucially, Partial Convolutions \cite{pconv-inpaint, pconv-pad} are used to prevent empty masked regions from affecting the output.

\subsection{Generated 3D Model Alignment}
\label{sec:supp_alignment}

Image-to-3D generators (such as TRELLIS2) typically output meshes in a canonical pose, e.g. always facing in the +z direction, and are agnostic to the true image camera position relative to the 3D object. Since our method assumes the pre-scan and the first dynamic frame to be aligned, as discussed in Sec.~\ref{sec:method-canonical}, we align and scale TRELLIS2 mesh to match the first video frame as follows. Given the initially generated 3D model, we start from placing virtual cameras on a hemisphere surrounding the model. The renders of the mesh on each camera are processed by VGGT \cite{wang2025vggt} along with the first video frame. We then sample the virtual camera with the most similar extrinsics to the first video frame estimated by VGGT. This serves as an initialization for our localization task. We then use MASt3R \cite{mast3r} to find 2D correspondences between the first video frame and the sampled virtual camera render. For each matched 2D point on the virtual render, we find its corresponding 3D point on the mesh by rendering the mesh depth on the camera and back-projecting. This yields 2D-to-3D correspondences from the first video frame to the 3D model points, and we use Perspective-n-Points algorithm (PnP) \cite{Fischler1981RandomSC, schoenberger2016sfm} to estimate the 6-DOF transformation of the mesh w.r.t. the video frame \cite{Khatib2025GeneralizableVL, liu2025gscpr}. Lastly, we align the mesh scale to the ViPE depth scale by the median depth ratio.

\subsection{Deduplication Masks}
\label{sec:supp_dedup}
As discussed in Sec.~\ref{sec:dmp}, we apply de-duplication masks on the output of \motiondipshort{} to ensure each 3D region is represented by a unique primitive across virtual views $\tilde{\cam}_{j}$. We start from back-projecting all Gaussians from a chosen view. Then, for each next view $j$, we render the depth and the transmittance of existing back-projected Gaussians to $\tilde{\cam}_{j}$. We then mark the pixels in $j$ "empty" if: (a) the rendered transmittance is below $0.1$, or (b) the difference between the rendered depth and $\tilde{\depth}_j$ is larger than a threshold $0.01$. This produces a set of de-duplication masks for each view $\tilde{\cam}_{j}$ depending on the starting view. In total, we use 8 virtual cameras, and get the masks for each possible starting view, obtaining 8 sets of de-duplication masks. We uniformly sample a single mask set for each training iteration.

\subsection{Hyperparameters}
\label{sec:supp_hyperparams}


We train our model using the Adam optimizer \cite{Kingma2014AdamAM} with a learning rate $2 \cdot 10^{-3}$ for 50,000 iterations with batch size 1. Training for a video with 150 frames takes approximately 10 hours on a single NVIDIA A100 GPU. In Eq.~\ref{eq:canonref}, Eq.~\ref{eq:lphoto} and Eq.~\ref{eq:lfinal}, we use the following weights in all our experiments: $\lambdassim=0.25, \lambdatv=0.1, \lambdatrack=1.0, \lambdadepth=100.0, \lambdareproj=100.0, \lambdacoarseiso=10.0, \lambdadenseiso=1.0, \lambdarigid=10.0 $. We set the spatial falloff parameter $\beta=2000$ for $\lossdenseiso$. In the coarse isometry loss, the number of nearest-neighbors in $\nncoarse$ is set to $0.01 \cdot | \pcoarse |$, while in the dense isometry loss, we use 200 nearest-neighbors in $\nn$.

\subsection{Truebones Benchmark}
\label{sec:supp_truebones}

For the Truebones benchmark, we select 7 animations of highly articulated animals from their dynamic mesh dataset, which are labeled as: "Bat/AttackBite" (60 frames), "Anaconda/Strike" (140 frames), "BrownBear/Rise" (91 frames), "Bear/WalkForward" (51 frames), "Coyote/Walking" (69 frames), "Camel/Restless" (139 frames), "Camel/Run" (61 frames). We create 150 pre-scan cameras around the object in the first frame, followed by cameras for the monocular dynamic sequence, which are smoothly moving around the object while being directed towards its center at every timestep. For test views, we place 4 static cameras facing each side of the object (front, back, left, right). All videos are rendered at 1024x1024 resoltion and 30fps. See our website for visualizations of training and testing videos.

\begin{figure}[h]
    \centering
    \includegraphics[width=0.5\columnwidth]{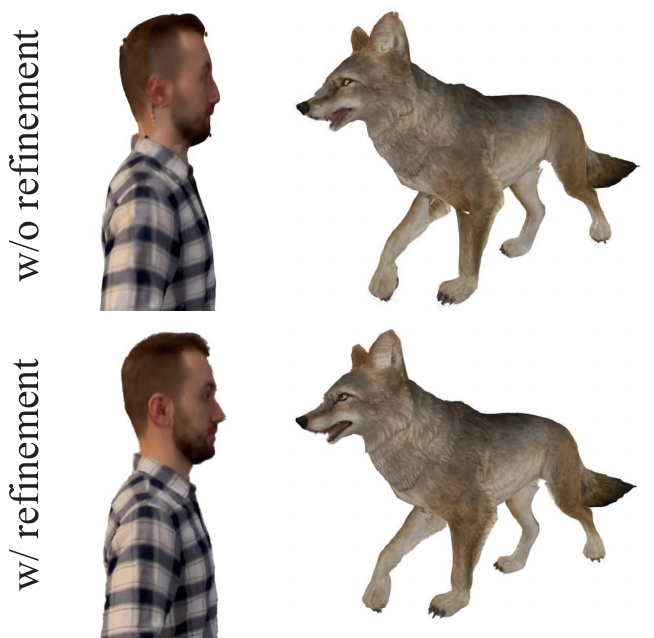}
    \caption{\emph{Pre-scan Refinement} optimization (Sec.~\ref{sec:method-canonical}) significantly improves high-frequency details in shape and texture, e.g. the geometry of the person's face, the coyote's fur and nose shape. Renders are obtained from test views.}
    \label{fig:supp_prescan_ref}
\end{figure}

\subsection{Tracking Evaluation Details}
\label{sec:supp_tracking_eval}
We evaluate 3D tracking performance in both TAP-Vid-3D and our Truebones benchmark by querying each track directly at its 3D query point and tracking both forward and backward in time. Specifically, given a set of dynamic Gaussians and a 3D query point, we sample the nearest Gaussian center to the query point, and output the Gaussian center at all frames as the long-range 3D track. To assess the complete tracking and geometric ability of the reconstruction models, we measure tracking accuracy in both occluded and visible frames.

For our Truebones 3D tracking benchmark, we take the 3D positions of all vertices for each dynamic mesh sequence as the ground-truth 3D tracks. In total, we extract 20,987 tracks from 7 animations. For all methods, we query and estimate each track at its first frame location.

\subsection{Canonical Depth Refinement}
\label{sec:supp_depth_ref}

As discussed in Sec.~\ref{sec:method-canonical}, the depth estimates of FoundationStereo serve as an excellent initialization for the canonical model, but lack high-frequency details. Fig. \ref{fig:supp_prescan_ref} shows the canonical Gaussians before and after the refinement optimization. As seen, the refinement optimization noticeably improves the high-frequency details in the shape and texture.

\end{document}